\documentclass[journal]{IEEEtran}

\ifCLASSINFOpdf
\else
   \usepackage[dvips]{graphicx}
\fi
\usepackage{url}
\usepackage{booktabs}
\usepackage{cite}
\usepackage{graphicx}
\usepackage{ifsym}

\usepackage{orcidlink}

\begin{document}

\title{LoASR-Bench: Evaluating Large Speech Language Models on Low-Resource Automatic Speech Recognition Across Language Families}

\author{Jianan Chen \orcidlink{0009-0006-2603-0074} \IEEEmembership{Member, IEEE}, Xiaoxue Gao$^*$ \IEEEmembership{Senior Member, IEEE}, Kawahara Tatsuya \IEEEmembership{Fellow, IEEE}, \\ and Nancy F. Chen \orcidlink{0000-0003-0872-5877} \IEEEmembership{Senior Member, IEEE}

% \thanks{The scripts as long with the fine-tuned checkpoints will be released to public once accepted.}

\thanks{Jianan Chen and Tatsuya Kawahara are with Kyoto University, Kyoto, Japan (e-mails: chen.jianan.32d@st.kyoto-u.ac.jp, kawahara@i.kyoto-u.ac.jp)}

\thanks{Xiaoxue Gao and Nancy F. Chen are with Institute for Infocomm Research, Agency for Science, Technology and Research, Singapore (e-mails: Gao\_Xiaoxue@a-star.edu.sg, nancy\_chen@a-star.edu.sg).}

\thanks{* Corresponding author.}
}

\markboth{Journal of \LaTeX\ Class Files, Vol. 14, No. 8, August 2015}
{Shell \MakeLowercase{\textit{et al.}}: Bare Demo of IEEEtran.cls for IEEE Journals}

\maketitle

\begin{abstract}

Large language models (LLMs) have driven substantial advances in speech language models (SpeechLMs), yielding strong performance in automatic speech recognition (ASR) under high-resource conditions.
However, existing benchmarks predominantly focus on high-resource languages, leaving the ASR behavior of SpeechLMs in low-resource languages insufficiently understood. 
This gap is critical, as practical ASR systems must reliably support low-resource languages and generalize across diverse language families, and it directly hinders the deployment of SpeechLM-based ASR in real-world multilingual scenarios. As a result, it is essential to evaluate SpeechLMs on low-resource languages to ensure their generalizability across different language families.
To address this problem, we propose \textbf{LoASR-Bench}, a comprehensive benchmark designed to evaluate \textbf{lo}w-resource \textbf{a}utomatic \textbf{s}peech \textbf{r}ecognition (\textbf{ASR}) of the latest SpeechLMs across diverse language families. LoASR-Bench comprises 25 languages from 9 language families, featuring both Latin and non-Latin scripts, enabling cross-linguistic and cross-script assessment of ASR performance of current SpeechLMs. Experimental results highlight the limitations of the latest SpeechLMs in handling real-world low-resource languages. 
\end{abstract}


\IEEEpeerreviewmaketitle

\section{Introduction}
The recent emergence of large language models (LLMs) has significantly propelled the development of speech language models (SpeechLMs)~\cite{an2024funaudiollmvoiceunderstandinggeneration,bai2024seedasrunderstandingdiversespeech, 10.1109/TASLP.2024.3379877}. These SpeechLMs typically leverage large-scale multimodal and multilingual pre-training to achieve strong performance across a wide range of speech-related tasks. With this rapid progress, benchmarks play a crucial role in systematically evaluating and comparing these SpeechLMs~\cite{chen2024voicebench, cui2024recent, 10.1093/nsr/nwae403}. Existing benchmarks primarily assess model performance on tasks such as language identification, automatic speech recognition (ASR), and speech translation. 

A common approach to applying LLMs to speech tasks is to construct cascaded systems, where speech is first transcribed with ASR models. The transcripts are then passed into LLMs for text-to-text generation~\cite{baevski2020wav2vec20frameworkselfsupervised, 10.1109/TASLP.2021.3122291, gao2024_icassp_cascaded}. In this work, we concentrate on ASR, aiming to provide a focused and in-depth evaluation of SpeechLMs in this foundational and widely deployed task. 

% Although the cascaded approach is practical for high-resource languages, performance tends to degrade in low-resource settings due to error propagation and limited training data~\cite{10848811,min2025end,shapira-etal-2025-measuring}. To address these limitations, recent advances in SpeechLMs have introduced end-to-end (E2E) architectures that are jointly trained on speech and text within a unified framework~\cite{ao-etal-2022-speecht5, Qwen-Audio}. Representative models include the Whisper~\cite{pmlr-v202-radford23a}, which leverages large-scale weakly supervised training, and the Qwen Audio family~\cite{chu2023qwenaudioadvancinguniversalaudio}, which integrates speech understanding with large language models. Additionally, self-supervised learning approaches such as XLSR-53~\cite{baevski2020wav2vec20frameworkselfsupervised} have shown promise in learning universal speech representations across languages.

Although cascaded approaches are practical for high-resource languages,  performance often degrades in low-resource settings due to error propagation and limited training data~\cite{10848811,min2025end,shapira-etal-2025-measuring}. To address these limitations, recent SpeechLMs have increasingly adopted end-to-end (E2E) architectures that are jointly trained on speech and text within a unified framework~\cite{zhang-etal-2022-speechut,ao-etal-2022-speecht5, chien-etal-2024-alignsspeechtext, liu24d_interspeech, Qwen-Audio, pmlr-v202-radford23a, chu2023qwenaudioadvancinguniversalaudio}. 
%They integrate speech understanding with large language models.
In addition, self-supervised learning approaches such as XLSR-53~\cite{baevski2020wav2vec20frameworkselfsupervised, radford2022whisper} have shown strong potential in learning universal speech representations across languages.

% SpeechLMs are increasingly trained in multilingual settings to facilitate cross-lingual knowledge transfer~\cite{inaguma2019multilingual, chou-etal-2023-toward, fang-etal-2024-llama-omni}. Such transfer is particularly beneficial for low-resource languages, where both speech and text data are scarce. By leveraging phonetic and semantic features across languages, SpeechLMs can improve ASR performance in low-resource settings~\cite{ruder-etal-2019-unsupervised, 10890363, 10.5555/3618408.3618627}. This multilingual training strategy enables models to generate universal representations and enhance cross-lingual generalization. 

SpeechLMs are increasingly trained in multilingual settings to facilitate cross-lingual knowledge transfer~\cite{inaguma2019multilingual, chou-etal-2023-toward, fang-etal-2024-llama-omni}. Such transfer is particularly beneficial for low-resource languages, where both speech and text data are limited. By leveraging shared phonetic and semantic features across languages, SpeechLMs can improve ASR performance in low-resource scenarios~\cite{ruder-etal-2019-unsupervised, 10890363, 10.5555/3618408.3618627}. This multilingual training strategy enables models to learn more universal representations.

% Consequently, numerous benchmarks have been developed to evaluate ASR systems across various aspects. Early benchmarks focused primarily on English and other high-resource languages. Recent efforts have expanded to multilingual settings, with benchmarks such as CommonVoice~\cite{ardila-etal-2020-common} and FLEURS~\cite{10023141} covering dozens of languages. However, these benchmarks often treat languages independently without systematic analysis of language family characteristics or script diversity. As a result, such evaluations provide limited insight into how ASR models generalize across languages within the same language family. This limitation hinders the understanding of multilingual ASR systems in low-resource settings.
% Several recent works have begun addressing low-resource ASR evaluation, including multilingual speech understanding~\cite{yang21c_interspeech}, region-specific language evaluations~\cite{shi23g_interspeech, shi24g_interspeech}, and language-focused ASR benchmarks~\cite{ko2024benchmarkingjapanesespeechrecognition}. Despite these contributions, existing benchmarks have critical limitations: they primarily focus on individual language performance without cross-family comparisons, lack systematic analysis of script-type effects (Latin vs. non-Latin), and often underrepresent typologically diverse language families such as Dravidian, Uralic, and Turkic.

Numerous benchmarks have been developed to evaluate ASR systems from different perspectives. Early benchmarks focused primarily on English and other high-resource languages\cite{kawahara2003benchmark, sehar-etal-2025-benchmarking, pratap2020massivelymultilingualasr50,yang21c_interspeech,shi23g_interspeech,shi24_mlsuperb2, ko2024benchmarkingjapanesespeechrecognition, zhao2025babelopenmultilinguallarge,9383459}. More recent efforts have expanded to multilingual settings, with benchmarks such as CommonVoice~\cite{ardila-etal-2020-common} and FLEURS~\cite{10023141}. However, these benchmarks typically treat languages independently and lack systematic analysis of typologically diverse language-family characteristics (i.e., Dravidian, Uralic, and Turkic) or script diversity (Latin vs. non-Latin). As a result, they provide limited insight into how ASR models generalize across languages.

\begin{figure*}
\centering
\includegraphics[width=1.5\columnwidth]{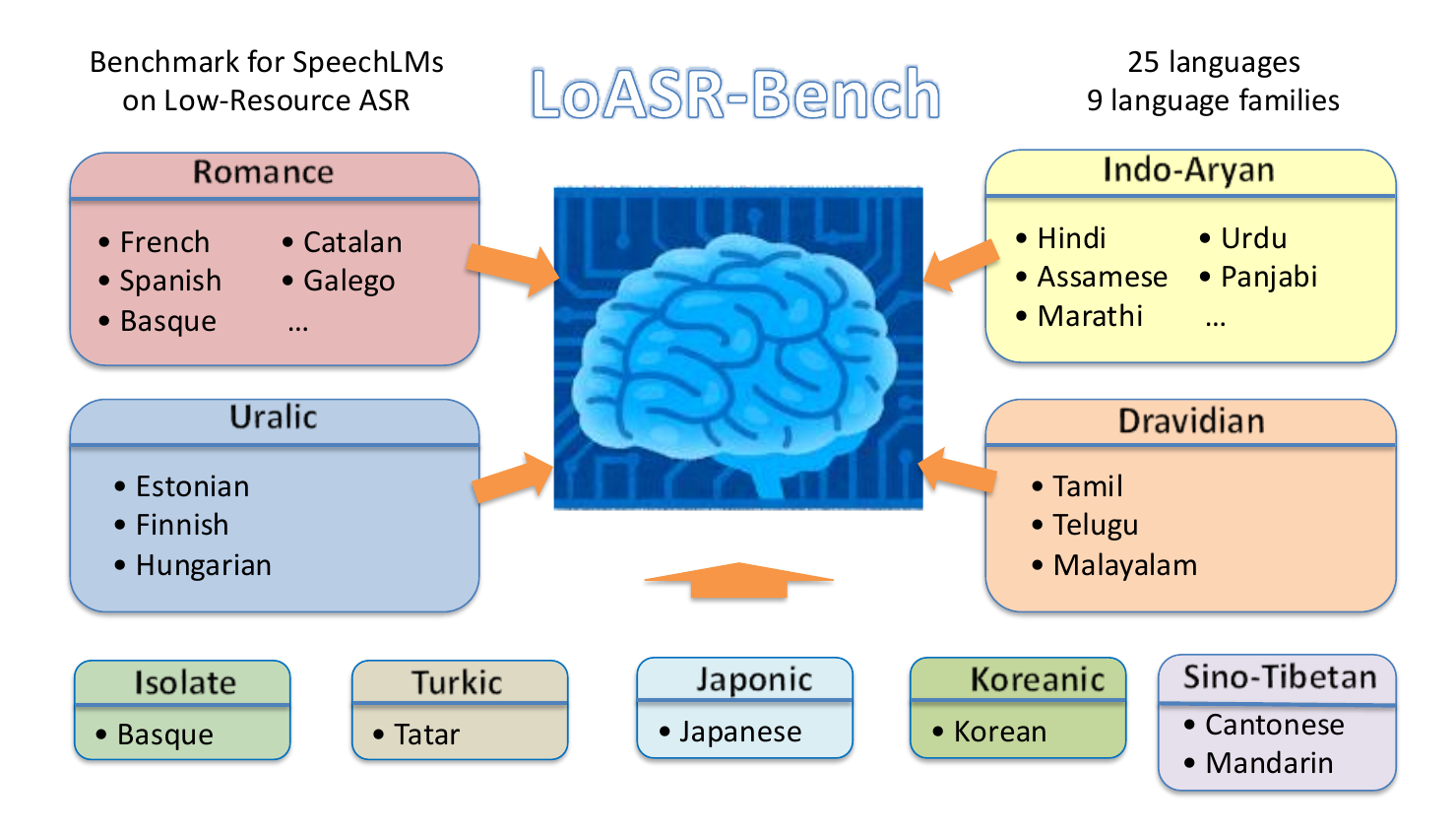}
\centering
\caption{Overview of LoASR-Bench covering low-resource languages from nine language families.}
\label{fig:illustration_benchmarking_across_families}
\end{figure*}

% Despite recent advances in ASR benchmarks~\cite{yang21c_interspeech, shi23g_interspeech, shi24g_interspeech, ko2024benchmarkingjapanesespeechrecognition}, critical limitations remain that hinder progress towards global applications of current SpeechLMs. First, these benchmarks primarily focus on high-resource languages, such as English, overlooking performance on low-resource languages. Second, most existing evaluations consider each language independently, without examining how linguistic structure and script type affect model performance~\cite{}. The extent to which these benefits transfer across typologically diverse language families and different writing systems remains underexplored. The structural and phonological relationships among languages, as well as the impact of writing systems (Latin vs. non-Latin scripts), have not been systematically studied. 

To address these problems, we introduce \textbf{LoASR-Bench}, a benchmark designed to provide a comprehensive evaluation of \textbf{Lo}w-Resource \textbf{A}utomatic \textbf{S}peech \textbf{R}ecognition. LoASR-Bench encompasses 25 languages across 9 typologically diverse language families (Dravidian, Romance, Uralic, Isolate, Japonic, Sino-Tibetan, Indo-Aryan, Turkic, and Koreanic), including both Latin and non-Latin scripts. To the best of our knowledge, \textbf{LoASR-Bench} is the first benchmark to systematically compare the performance of SpeechLMs across different language families and script types. By introducing diversity across language families and scripts, \textbf{LoASR-Bench} enables a more realistic evaluation of SpeechLMs' performance on low-resource languages. We evaluate state-of-the-art SpeechLMs, including XLSR-53, Whisper, and Qwen models, and demonstrate that current models encounter unique challenges in low-resource languages that are absent in resource-rich ones. Consequently, these gaps risk overestimating the generalizability of SpeechLMs to real-world low-resource languages.  We find that models perform substantially better on Latin-script languages (i.e., Romance family) compared to non-Latin scripts (i.e., Dravidian and Indo-Aryan families). Notably, we observe that model size and multilingual pre-training significantly impact performance. 

% Our contributions are:
% \begin{itemize}
%     \item We present \textbf{LoASR-Bench}, a new benchmark designed specifically for low-resource ASR across diverse language families and scripts, covering 25 languages from 9 typologically distinct families.
%     \item We conduct a systematic evaluation of different SpeechLMs, revealing significant performance variations and overall trends on low-resource languages.
%     \item We perform comprehensive cross-family and cross-script comparisons, demonstrating that Latin-script languages consistently achieve lower error rates than non-Latin-script languages and highlighting how language-family characteristics influence model performance.
%     \item We offer both quantitative and qualitative analyses that uncover current challenges faced by SpeechLMs. We find that Latin-script languages consistently achieve lower error rates than non-Latin-script languages. We then identify future directions for improvement. 
%     \item We have fine-tuned the XLSR-53, Whisper, and Qwen models across 25 target languages, including some low-resource languages without publicly available checkpoints. To support reproducibility, the codes and fine-tuned checkpoints will be made publicly available upon acceptance.
% \end{itemize}
Our main contributions include:
\begin{itemize}
    \item We present \textbf{LoASR-Bench}, a new benchmark designed specifically for low-resource ASR across diverse language families, covering 25 languages from 9  language families.
    \item We conduct a systematic evaluation of different SpeechLMs, revealing significant performance variations on low-resource languages.
    \item We perform comprehensive cross-family and cross-script comparisons, showing that languages with Latin scripts achieve lower error rates than non-Latin languages.
    \item We conduct both language-aware and language-unaware inference for SpeechLMs and observe that identifying languages is crucial for ASR performance.
    \item We propose to fine-tune XLSR-53, Whisper, and Qwen models across 25 target languages, covering several low-resource languages that lack publicly available checkpoints. This fine-tuning effort establishes a unified and reproducible set of SpeechLMs for low-resource ASR, with training code and checkpoints to be released upon acceptance.
\end{itemize}
\section{LoASR-Bench}
To address the lack of systematic evaluation for low-resource ASR, we introduce \textbf{LoASR-Bench}, a comprehensive multilingual benchmark designed to evaluate SpeechLMs across 25 low-resource languages spanning 9 language families, as illustrated in Fig.~\ref{fig:illustration_benchmarking_across_families}.

\begin{table*}[ht]
\caption{ASR performance across language families and models. CER\% is reported for Latin-script languages, and WER\% for non-Latin scripts. Train (h) denotes training data duration in hours. Best results for each language or average row are shown in \textbf{bold}. Lower is better.}
\centering
% \small
% \normalsize
\scriptsize
\renewcommand{\arraystretch}{1.1}
\setlength{\tabcolsep}{2.8pt}
\begin{tabular}{|l|l|r|c|c|c|c|c|c|}
\hline
\textbf{Language Family} & \textbf{Language (ISO)} & \textbf{Train (h)} &
\textbf{xlsr-53} & \textbf{Whisper-M} & \textbf{Whisper-L} &
\textbf{Qwen2-Audio} & \textbf{Qwen2-Audio ft} & \textbf{Qwen3-Omni} \\
\hline\hline

% ---------------- Dravidian ----------------
\multicolumn{9}{|l|}{\textit{Dravidian (WER)}: non-Latin script} \\ \hline
 & Tamil (ta)      & 233 & 0.36 & 0.52 & 0.48 & 3.05 & 0.63 & \textbf{0.31} \\
 & Telugu (te)     & 1   & 1.00 & \textbf{0.33} & 0.34 & 4.80 & 1.13 & 0.71 \\
 & Malayalam (ml)  & 1   & \textbf{0.36} & 0.56 & 0.49 & 4.73 & 0.46 & 0.85 \\
 & \textbf{Avg.}   & --  & 0.57 & 0.47 & \textbf{0.44} & 4.19 & 0.74 & 0.62 \\
\hline\hline

% ---------------- Romance ----------------
\multicolumn{9}{|l|}{\textit{Romance (CER)}: Latin script} \\ \hline
 & Portuguese (pt) & 208 & 0.21 & 0.21 & 1.86 & 3.26 & 0.26 & \textbf{0.07} \\
 & Spanish (es)    & 549 & 0.05 & 0.04 & 0.04 & 2.20 & 0.04 & \textbf{0.03} \\
 & French (fr)     & 989 & 0.11 & 0.21 & 0.20 & 2.60 & 0.21 & \textbf{0.06} \\
 & Italian (it)    & 353 & 0.25 & 0.18 & 0.20 & 0.19 & 0.30 & \textbf{0.06} \\
 & Catalan (ca)    & 2670& \textbf{0.07} & 0.17 & 0.15 & 2.40 & 0.10 & 0.17 \\
 & Galician (gl)   & 48  & 0.08 & 0.15 & 0.95 & 1.75 & \textbf{0.03} & 0.10 \\
 & Romanian (ro)   & 20  & 0.48 & 0.82 & 0.76 & 3.31 & 0.81 & \textbf{0.12} \\
 & \textbf{Avg.}   & --  & 0.18 & 0.25 & 0.59 & 2.24 & 0.25 & \textbf{0.09} \\
\hline\hline

% ---------------- Uralic ----------------
\multicolumn{9}{|l|}{\textit{Uralic (CER)}: Latin script} \\ \hline
 & Estonian (et)   & 46 & 0.24 & \textbf{0.12} & 0.14 & 0.75 & 0.43 & 0.34 \\
 & Finnish (fi)    & 12 & 0.25 & \textbf{0.17} & 0.19 & 1.22 & 0.90 & 0.86 \\
 & Hungarian (hu)  & 92 & 0.12 & 0.09 & 0.20 & 2.30 & 0.57 & \textbf{0.08} \\
 & \textbf{Avg.}   & -- & 0.20 & \textbf{0.13} & 0.18 & 1.42 & 0.63 & 0.43 \\
\hline\hline

% ---------------- Isolate ----------------
\multicolumn{9}{|l|}{\textit{Isolate (CER)}: Latin script} \\ \hline
 & Basque (eu)     & 220 & 0.45 & 0.06 & \textbf{0.04} & 3.55 & 0.06 & 0.25 \\
 & \textbf{Avg.}   & --  & 0.45 & 0.06 & \textbf{0.04} & 3.55 & 0.06 & 0.25 \\
\hline\hline

% ---------------- Japonic ----------------
\multicolumn{9}{|l|}{\textit{Japonic (WER)}: non-Latin script} \\ \hline
 & Japanese (ja)   & 122 & 0.56 & 0.41 & 0.44 & 0.35 & \textbf{0.21} & \textbf{0.21} \\
 & \textbf{Avg.}   & --  & 0.56 & 0.41 & 0.44 & 0.35 & \textbf{0.21} & \textbf{0.21} \\
\hline\hline

% ---------------- Sino-Tibetan ----------------
\multicolumn{9}{|l|}{\textit{Sino-Tibetan (WER)}: non-Latin script} \\ \hline
 & Cantonese (yue) & 23 & 0.33 & 0.33 & 0.35 & 17.42 & \textbf{0.01} & 0.15 \\
 & Chinese (zh-CN) & 12 & 0.54 & 0.45 & 0.48 & 2.39 & 0.45 & \textbf{0.09} \\
 & \textbf{Avg.}   & -- & 0.44 & 0.39 & 0.42 & 9.91 & 0.23 & \textbf{0.12} \\
\hline\hline

% ---------------- Indo-Aryan ----------------
\multicolumn{9}{|l|}{\textit{Indo-Aryan (WER)}: non-Latin script} \\ \hline
 & Hindi (hi)      & 14 & 0.57 & 1.53 & 0.81 & 3.14 & \textbf{0.06} & 0.11 \\
 & Urdu (ur)       & 64 & 0.99 & 0.64 & 0.64 & 3.80 & 0.89 & \textbf{0.63} \\
 & Assamese (as)   & 2.7& 0.98 & 0.25 & 0.26 & 3.42 & \textbf{0.08} & 0.57 \\
 & Marathi (mr)    & 19 & 0.63 & 0.30 & 0.29 & 2.64 & \textbf{0.17} & 0.29 \\
 & Bengali (bn)    & 54 & 0.57 & 0.34 & 0.33 & 2.67 & 0.24 & \textbf{0.12} \\
 & Punjabi (pa)    & 4  & 0.98 & 0.63 & 0.59 & 2.31 & \textbf{0.09} & 0.77 \\
 & \textbf{Avg.}   & -- & 0.79 & 0.62 & 0.49 & 2.99 & \textbf{0.26} & 0.42 \\
\hline\hline

% ---------------- Turkic ----------------
\multicolumn{9}{|l|}{\textit{Turkic (CER)}: Latin script} \\ \hline
 & Tatar (tt)      & 31 & 0.75 & 0.46 & 0.34 & 0.98 & \textbf{0.31} & 0.85 \\
 & \textbf{Avg.}   & -- & 0.75 & 0.46 & 0.34 & 0.98 & \textbf{0.31} & 0.85 \\
\hline\hline

% ---------------- Koreanic ----------------
\multicolumn{9}{|l|}{\textit{Koreanic (WER)}: non-Latin script} \\ \hline
 & Korean (ko)     & 2 & 0.20 & 0.18 & 0.17 & 0.34 & 0.12 & \textbf{0.09} \\
 & \textbf{Avg.}   & -- & 0.20 & 0.18 & 0.17 & 0.34 & 0.12 & \textbf{0.09} \\
\hline\hline

% ---------------- Script-level averages ----------------
\textbf{Latin Avg.} & -- & -- & 0.27 & \textbf{0.23} & 0.42 & 2.03 & 0.33 & 0.26 \\
\hline
\textbf{non-Latin Avg.} & -- & -- & 0.62 & 0.50 & 0.44 & 3.93 & \textbf{0.35} & 0.38 \\
\hline\hline
\textbf{Total Avg.} & -- & -- & 0.45 & 0.37 & 0.43 & 3.02 & 0.34 & \textbf{0.32} \\
\hline
\end{tabular}
\label{tab:evaluation_results_merged}
\end{table*}

%%==================== Table instruction ====================
\begin{table}[t]
\caption{Inferences instructions used for Qwen models in \textbf{LoASR-Bench}. 
Language-Aware prompts include the target language name explicitly.}
\centering
\small
\begin{tabular}{|l|l|}
\hline
\textbf{Model} & \textbf{Instruction} \\
\hline
Language Unaware & Transcribe the audio into text. \\
Language Aware& Transcribe the [Language Name] \\ 
 & audio into text. \\
\hline
\end{tabular}
\label{tab:qwen_instructions}
\end{table}
%%==================== Table instruction End ====================

\subsection{Language Families}

Motivated by the absence of systematic benchmarks that assess SpeechLM in typologically diverse language families, \textbf{LoASR-Bench} incorporates cross-family evaluations to assess model generalization and robustness under typological diversity. Evaluating languages from diverse linguistic families is essential, as variations in phonology, morphology, syntax, and writing systems can substantially influence ASR performance, particularly in low-resource settings.

Our \textbf{LoASR-Bench} includes nine major language families, covering the Dravidian family (Tamil, Telugu, Malayalam), Romance (Portuguese, Spanish, French, Italian, Catalan, Galician, Romanian), Uralic (Estonian, Finnish, Hungarian),  Isolate (Basque), Japonic (Japanese), Sino-Tibetan (Cantonese and Mandarin Chinese), Indo-Aryan (Hindi, Urdu, Assamese, Marathi, Bengali, Punjabi), Turkic (Tatar), and Koreanic (Korean). This broad coverage enables fine-grained analysis of SpeechLM behavior across diverse linguistic families and scripts, revealing strengths and limitations that are not observable in single-language or region-specific benchmarks.

\subsection{SpeechLMs and Low-resource Language Fine-Tuning}
We evaluate a diverse set of widely adopted and state-of-the-art SpeechLMs that differ in architecture, scale, and training paradigm. These include XLSR-53, a self-supervised encoder model pre-trained on 53 languages based on wav2vec 2.0~\cite{baevski2020wav2vec20frameworkselfsupervised}; Whisper-Medium and Whisper-Large-v3, encoder–decoder ASR models trained on 680k hours of weakly supervised multilingual data~\cite{radford2022whisper}; and multimodal SpeechLMs from the Qwen family, including Qwen2-Audio-7B and Qwen3-Omni-30B~\cite{chu2024qwen2,xu2025qwen2,yang2025qwen3}. Notably, Qwen3-Omni further extends multilingual capability with data covering over 100 languages.

To support systematic and fair evaluation in low-resource settings, we propose to fine-tune all evaluated SpeechLMs across 25 target languages using Common Voice 16.1~\cite{ardila-etal-2020-common} under a unified and scalable training protocol.
Each model is trained for 10 epochs with a batch size of 8, a learning rate of 5e-5, and mixed-precision training (fp16). 
Model selection adopts script-aware validation, using CER for Latin-script languages and WER for non-Latin scripts, with the best-performing checkpoint retained for evaluation. Text normalization is applied consistently across all models and languages to ensure comparability. This proposed large-scale fine-tuning strategy establishes a controlled and reproducible benchmark setting, enabling direct comparison across languages, scripts, and language families. All experiments are implemented using the official Hugging Face Transformers framework.

We conduct the experiments in a language-unaware setting shown in Table~\ref{tab:qwen_instructions}, where the language name is omitted from the inference instruction to isolate the effect of explicit language cues. For comparison, we also evaluate a language-aware setting on Qwen3-Omni. 

% In this study, we present \textbf{LoASR-Bench}, a new benchmark designed specifically for low-resource ASR across diverse language families and scripts, covering 25 languages from 9 language families. We conduct comprehensive cross-family and cross-script evaluation of different SpeechLMs, revealing significant performance variations on these languages. We find Latin-script languages consistently achieve lower error rates than non-Latin-script languages. To support reproducibility, the codes and fine-tuned checkpoints will be made publicly available upon acceptance. 

% Our work addresses these gaps by introducing LoASR-Bench, which explicitly evaluates SpeechLMs across nine language families with diverse typological characteristics and writing systems. Unlike prior benchmarks that treat languages as independent entities, LoASR-Bench enables systematic comparison of how language family membership and script type influence model performance, providing insights into the true multilingual capabilities of current SpeechLMs.

\begin{table*}[ht]
\caption{Comparison of language-unaware and language-aware inference.}
\centering
\scriptsize
\renewcommand{\arraystretch}{1.0}
\setlength{\tabcolsep}{2.6pt}
\begin{tabular}{|l|ccc|ccccccc|ccc|c|c|cccccc|c|c|}
\hline
 & \multicolumn{3}{c|}{\textit{Dravidian (CER)}} 
 & \multicolumn{7}{c|}{\textit{Romance (WER)}}
 & \multicolumn{3}{c|}{\textit{Uralic (WER)}}
 & \textit{Isolate} 
 & \textit{Japonic}
 & \multicolumn{6}{c|}{\textit{Indo-Aryan (CER)}}
 & \textit{Sino-Tibetan} 
 & \textbf{Avg.} \\
\textbf{Model} 
& \textbf{ta} & \textbf{te} & \textbf{ml}
& \textbf{pt} & \textbf{es} & \textbf{fr} & \textbf{it} & \textbf{ca} & \textbf{gl} & \textbf{ro}
& \textbf{et} & \textbf{fi} & \textbf{hu}
& \textbf{eu}
& \textbf{ja}
& \textbf{hi} & \textbf{ur} & \textbf{as} & \textbf{mr} & \textbf{bn} & \textbf{pa}
& \textbf{yue}
& \textbf{} \\
\hline
Unaware
& 0.31 & 0.71 & 0.85 
& 0.07 & 0.03 & 0.06 & 0.06 & 0.17 & 0.10 & 0.12
& 0.34 & 0.86 & 0.08
& 0.25
& 0.21
& 0.11 & 0.63 & 0.57 & 0.29 & 0.12 & 0.77
& 0.15
& 0.29 \\
Aware
& 0.53 & 0.19 & 0.38 
& 0.09 & 0.06 & 0.14 & 0.11 & 0.33 & 0.11 & 0.19
& 0.66 & 0.25 & 0.57
& 0.69
& 0.19
& 0.05 & 0.19 & 0.32 & 0.12 & 0.10 & 0.20
& 0.13
& \textbf{0.27} \\
\hline
\end{tabular}
\label{tab:unaware_vs_langaware}
\end{table*}

% Following standard practice in multilingual ASR evaluation, we report WER for Latin-script languages (Romance, Uralic, Turkic, and Isolate families) and CER for non-Latin scripts (Dravidian, Indo-Aryan, Sino-Tibetan, Japonic, and Koreanic families). 
% We conduct the experiments in a language-unaware setting, where the language name is omitted from the inference instruction to isolate the effect of explicit languages.

\section{Results and Analysis}

\textbf{XLSR-53 vs. Whisper vs. Qwen:} As Table~\ref{tab:evaluation_results_merged} shows, the XLSR-53 model demonstrates strong performance, particularly for Romance and Uralic languages. While Whisper models demonstrate competitive performance across most languages. However, Whisper-L showing less comparable results to Whisper-M in some cases, which might be caused by the mismatch of data size and model parameters. After fine-tuning Qwen2-Audio and with the latest Qwen3-Omni model, the Qwen family consistently outperforms Whisper baselines. However, for certain languages such as Tamil, Qwen2-Audio shows higher error rates compared to Whisper. Meanwhile, Qwen3-Omni does not fully outperform Qwen2-Audio across all languages, especially languages in Indo-Aryan and Turkic family, which are non-Latin script languages.

% Qwen3-Omni achieves the best results across nearly all languages, while the base Qwen2-Audio (without fine-tuning) performs significantly worse, likely due to limited language coverage in its training data. Fine-tuning Qwen2-Audio (Qwen2-Audio ft) substantially improves performance, demonstrating the importance of language-specific adaptation to the base models. These findings highlight the advantages of both multilingual pre-training and fine-tuning for ASR, with Qwen3-Omni benefiting from exposure to over 100 languages, demonstrating the importance of language data.

\textbf{Comparison Across Language Families:} Table~\ref{tab:evaluation_results_merged} presents averaged error rates after each language family, revealing notable variations. Romance languages show relatively low error rates across all models, particularly with Qwen3-Omni for most languages. Dravidian languages, however, present significant challenges for the Qwen2-Audio base model but show dramatic improvements with fine-tuning. Furthermore, almost all language families can benefit from language-specific fine-tuning. These findings suggest that lower-resource language families like Dravidian and Indo-Aryan require language-specific fine-tuning and larger multilingual models to achieve comparable performance.

\textbf{Latin and non-Latin Scripts:} As shown in Table~\ref{tab:evaluation_results_merged}, a clear performance gap exists between languages written in Latin and non-Latin scripts. Latin script languages show averaged lower error rate across all models, whereas non-Latin scripts generally yield averaged higher error rate. This finding suggests that script-related features (such as morphology) plays an important role in ASR performance for low-resource languages.

\begin{figure}
\centerline{\includegraphics[width=0.9\columnwidth]{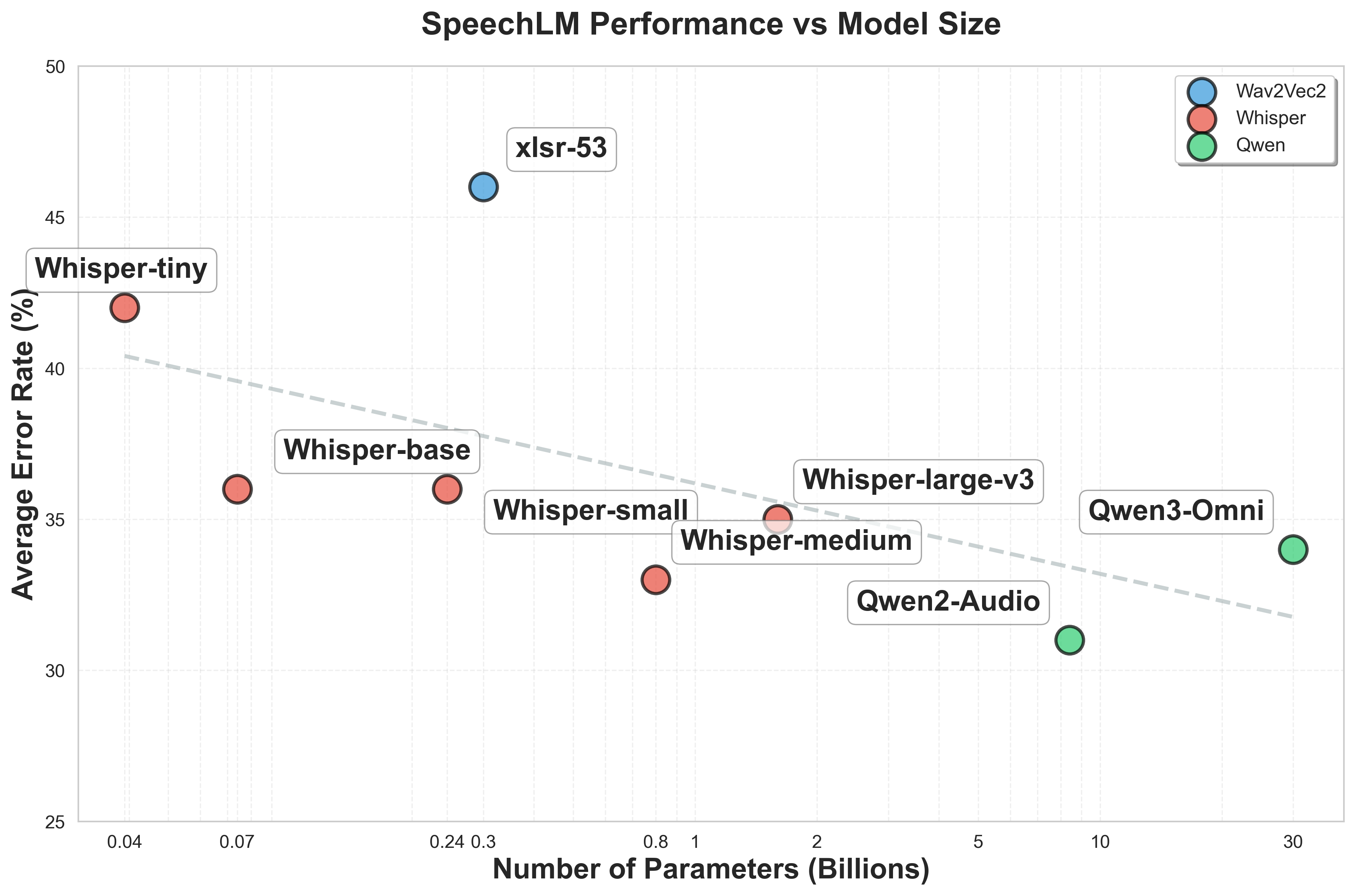}}
\caption{Average error rates across all languages vs. model size.}
\label{fig:model_size_vs_performance}
\end{figure}

\textbf{Model Size vs. Performance:}
As shown in Fig.~\ref{fig:model_size_vs_performance}, we can see a negative correlation between model parameters and error rates, indicating that larger models tend to perform better. However, although the later SpeechLM models are not solely designed for the ASR task, the detailed results show that scaling from 0.04B to 30B parameters only reduces error rates from about 0.45 to 0.32. This indicates that raw model size alone provides minimal additional benefit for low-resource multilingual ASR on our benchmark. 

Meanwhile, despite the smaller models showing performance degradation compared to the bigger models, the gap remains relatively modest, suggesting that for scenarios where computing resources are constrained, smaller models offer a reasonable balance between computational costs and ASR accuracy for low-resource languages.

\textbf{Language-Unaware vs. Language-Aware Inference:} To examine the impact of language-specific prompting on SpeechLMs, we report both language-aware and language-unaware inference results in Table~\ref{tab:unaware_vs_langaware}. The experiments were conducted using Qwen3-Omni, with and without the explicit language name in the instructions in Table~\ref{tab:qwen_instructions}. The results show that the language-aware setting outperforms the language-unaware one on average, yielding lower error rates across most languages, particularly within the Indo-Aryan family. Notably, substantial improvements are observed for several languages, such as Telugu (te), Finnish (fi), and Punjabi (pa). Despite the improvements, we can also observe performance degradation in some languages, such as Tamil (ta) and others.

These findings suggest that explicitly providing the language name during inference is crucial for achieving optimal ASR performance. However, in real-world scenarios, the spoken language may be unknown. This highlights an important direction for future work—integrating language identification (LID) with ASR within SpeechLMs to enable fully autonomous multilingual inference.

\vspace{-0.2cm}
\section{Conclusion}
% We introduce LoASR-Bench, a benchmark for evaluating large speech–language models on Low-Resource ASR across 25 languages in 8 language families. Our results highlight both the strengths and the limitations of the latest SpeechLM models. The state-of-the-art Qwen3-Omni 30B exhibits significant gains compared to earlier versions and the Whisper baselines, confirming the value of diversity in multilingual training data. 
% However, significant gaps remain for languages that use non-Latin and Latin scripts. 
% Future work will integrate automatic language identification into E2E ASR. Our codes will be released once accepted.
% % Future work should focus on integrating automatic language identification into end-to-end ASR with SpeechLMs and selecting the correct script for decoding non-Latin, low-resource languages. 

We introduce LoASR-Bench, a benchmark evaluating speech language models on low-resource ASR across 25 languages from 9 typologically diverse language families. Our evaluation reveals that Qwen3-Omni demonstrates substantial improvements over earlier models and Whisper baselines, confirming the value of extensive multilingual pretraining. Furthermore, target-language fine-tuning plays an important role, with fine-tuned Qwen2-Audio showing dramatic error reductions compared to the base model without fine-tuning. 

We find that despite the success of the current SpeechLMs, significant challenges remain. Firstly, performance gaps persist between Latin and non-Latin scripts, with frequent script confusion and language misidentification. Languages with extremely limited data ($<$1 hour) show substantially higher error rates, highlighting the remaining data scarcity issues for both pre-training and fine-tuning, which indicate a deployment bottleneck for real-world low-resource languages. We also find that identifying the language is crucial in the inference of ASR.

\end{document}